# An Experimental Comparison of Several Clustering and Initialization Methods

**Marina Meilă**
mmp@ai.mit.edu
Center for Biological and
Computational Learning
Massachusetts Institute of Technology
Cambridge, MA 02142

**David Heckerman**
heckerma@microsoft.com
Microsoft Research
Redmond WA 98052-6399

## Abstract

We examine methods for clustering in high dimensions. In the first part of the paper, we perform an experimental comparison between three batch clustering algorithms: the Expectation–Maximization (EM) algorithm, a "winner take all" version of the EM algorithm reminiscent of the K-means algorithm, and model-based hierarchical agglomerative clustering. We learn naive-Bayes models with a hidden root node, using high-dimensional discrete-variable data sets (both real and synthetic). We find that the EM algorithm significantly outperforms the other methods, and proceed to investigate the effect of various initialization schemes on the final solution produced by the EM algorithm. The initializations that we consider are (1) parameters sampled from an uninformative prior, (2) random perturbations of the marginal distribution of the data, and (3) the output of hierarchical agglomerative clustering. Although the methods are substantially different, they lead to learned models that are strikingly similar in quality.

## 1 Introduction

The present work concentrates on the unsupervised learning of a simple but widely used mixture model of factored distributions: a *naive-Bayes model* with a hidden root node (e.g., Clogg, 1995; Cheeseman & Stutz, 1995). This learning task is often referred to as *clustering*. The conceptual simplicity of the naive-Bayes model and its ease of implementation make it an ideal prototype and benchmark model. Moreover, in higher dimensions, where both the search for model structure and escaping local optima in the parameter space can substantially slow computation, the relatively small number of parameters of the naive-Bayes model make it more appealing than other complex models.

Nonetheless, even for this relatively simple model, high dimensional domains (tens to hundreds of variables) can present a challenge for both structure and parameter search algorithms. The aim of our work is to study the behavior of a number of commonly used algorithms for learning the parameters of mixture models on data of high dimensionality. Within this framework, special attention will be given to the initialization issue: How important is the choice of the initial parameters of an iterative algorithm and how do we find a good set of initial parameters?

In Section 2, we introduce the clustering problem and the learning algorithms that we shall compare. All algorithms are batch algorithms as opposed to on-line ones. They are Expectation-Maximization (EM), Classification EM (CEM)—a "winner take all" version of the EM algorithm reminiscent of the K-means algorithm—and hierarchical agglomerative clustering (HAC). In Sections 3 and 4, we describe our experimental procedure and datasets, respectively. In Section 5, we compare the three learning algorithms. The experiments suggest that the EM algorithm is the best method under a variety of performance measures. Consequently, in Section 6, we study the initialization problem for EM by comparing three different initialization schemes. Finally, in Section 7, we draw conclusions and point to various directions for further research. Details about HAC and its implementation are presented in Meilă & Heckerman (1998).

## 2 Model, algorithms, and performance criteria

### 2.1 The clustering problem

Here we describe the clustering model and formulate the learning problem that the algorithms under study will attempt to solve. Assume the do-main of interest is described by the vector variable $X = (X_1, X_2, \ldots X_n)$. (We follow the usual convention of denoting random variables with upper case letters and their states with lower case letters.) The clustering model for the variables $X$ is given by

$$P(X) = \sum_{k=1}^{K} \lambda_k \prod_{i=1}^{n} P(X_i | \text{class} = k) \quad (1)$$

$$\sum_{k=1}^{K} \lambda_k = 1, \qquad \lambda_k \geq 0.$$

In general, each variable $X_1, X_2, \ldots X_n$ may be continuous or discrete. Throughout this paper, however, we assume that all the variables are discrete and that $P(X_i|\text{class} = k)$ are multinomial distributions. This model is sometimes referred to as a *multinomial-mixture model*. The model can also be viewed as a naive-Bayes model in which a hidden variable "class" renders the variables $X_1, \ldots, X_n$ mutually independent. Under this interpretation, the values $\lambda_k, k = 1, \ldots K$ correspond to the probability distribution of the variable "class".

Having a database of $N$ observations or *cases* $D = \{x^1, x^2, \ldots x^N\}$ the clustering problem consists of finding the model—the model structure (i.e., number of classes) and parameters for that structure—that best fits the data $D$ according to some criterion.

In what follows, we sometimes refer to the distribution $P(x|\text{class} = k)$ as the *class* or *cluster k*. Also, we sometimes refer to a case for which the variable class is set to $k$ as being *assigned* to class $k$, and sometimes refer to the set of all cases having this property as the *class* or *cluster k*.

### 2.2 Clustering algorithms

The clustering algorithms that we consider choose the best model in two stages. First, a Bayesian criterion is used to choose the best model structure. Then, the parameters for the best model structure are chosen to be either those parameters whose posterior probability given data is a local maximum (MAP parameters) or those parameters whose corresponding likelihood is a local maximum (ML parameters).

The Bayesian criterion for selecting model structure that we use is the log posterior probability of model structure given the training data $\log P(K|D_{train})$. (We use $K$ to denote a model structure with $K$ classes.) Because we assume uniform model-structure priors, this criterion reduces to the *log marginal likelihood* of the model structure $\log P(D_{train}|K)$. In our experiments, we approximate this criterion using the method of Cheeseman and Stutz (1995) (see also Chickering & Heckerman, 1997). Figure 1 shows how the log marginal likelihood varies as a function of $K$ under one particular experimental condition. Curves for other experimental conditions are qualitatively similar. As $K$ increases, the marginal likelihood first increases and then decreases. The decrease occurs because the Bayesian criterion has a built-in penalty for complexity.

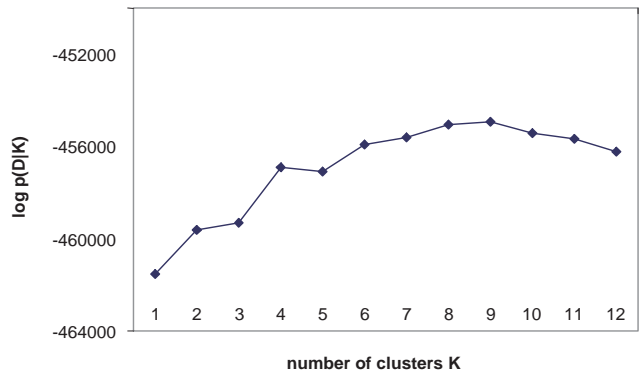

Figure 1: Log marginal likelihood as a function of cluster size K for a typical run.

We examine several algorithms for learning the parameters of a given model structure: the Expectation-Maximization (EM) algorithm (Dempster, Laird, & Rubin, 1977), the Classification EM (CEM) algorithm (Celeux & Govaert, 1992), and model-based hierarchical agglomerative clustering (HAC) (e.g., Banfield & Raftery, 1993). Sometimes, we shall refer to these parameter-learning algorithms simply as clustering algorithms.

The EM algorithm is iterative, consisting of two alternating steps: the Expectation (E) step and the Maximization (M) step. In the E step, for every $x^j$, we use the current parameter values in the model to evaluate the posterior distribution of the hidden class node given $x^j$. We then assign the case fractionally to each cluster according to this distribution. In the M step, we reestimate the parameters to be the MAP (or ML) values given this fractional assignment. The EM algorithm finds a local maximum for the parameters (MAP or ML). The maximum is local in the sense that any perturbation of the parameters would decrease the parameter posterior probability or data likelihood. The algorithm finds the maximum to any desired (non-perfect) precision in a finite number of steps.

The CEM algorithm is similar to the EM algorithm in that CEM also has E and M steps. The algorithm differs from EM in that, within the E step, a case $x^j$ is assigned fully to the class $k$ that has the highest posterior probability given $x^j$ and the current parameter values. That is, no fractional assignments are permitted. The M step in the CEM algorithm is identical to that in the EM algorithm. For continuous variable domains in which the mixture components are spherical Gaussian distributions, the CEM algorithm for ML parameters is equivalent to the K-means algorithm. The CEM algorithm converges to local MAP (or ML) parameters under the constraint that each case is assigned to only one cluster. Unlike EM, the CEM algorithm converges completely in a finite number of steps.

The EM and CEM algorithms require initial parameter values. We consider various initialization methods in the following section.

Hierarchical agglomerative clustering is substantially different from EM or CEM. When using hierarchical agglomerative clustering, we construct $K$ clusters from a larger number of smaller clusters by recursively merging the two clusters that are closest together. We start with $N$ clusters, each containing one case $x^j$. Each merge reduces the number of clusters by one. The algorithm stops when the desired number of clusters $K$ is reached. One essential ingredient of the algorithm is, of course, the intercluster "distance"[1] $d(k, l)$. The particular form of HAC that we examine is model based in the sense that the distance used for agglomeration is derived from a probabilistic model. Banfield & Raftery (1993) introduced a distance $d(k, l)$ derived from a Gaussian mixture model. This distance is equal to the decrease in likelihood resulting by the merge of clusters $k$ and $l$. Fraley (1997) derives this distance measure for special cases of Gaussian models, and describes algorithms for accomplishing the clustering in time and memory proportional to $N^2$. Here we derive this distance metric for mixtures of independent multinomially distributed variables (i.e., discrete naive-Bayes models). (The distance metric can be extended in a straightforward manner to consider differences in parameter posterior probability.) The algorithm we use requires memory only linear in $N$, and its running time is typically quadratic in $N$. With minor modifications, the same algorithm can handle certain classes of Gaussian mixtures.

Consider the likelihood of the data $D$ given an assignment of the cases to clusters:

$$L(D|C_1, \ldots, C_K) = \sum_{k=1}^{K} \sum_{x^j \in C_k} \log P(x^j|k) \quad (2)$$
$$= \sum_{k=1}^{K} L_k(\theta_k)$$

---
[1]This measure is positive and symmetric, but may not always satisfy the triangle inequality.

where $L_k(\theta_k)$ denotes the contribution of cluster $C_k$ to the log-likelihood, and $\theta_k$ represents the set of ML parameters of the distribution $P(x|k)$. Note that the term $L_k(\theta_k)$ depends only on the cases in cluster $C_k$. By merging clusters $k$ and $l$ and assigning all their cases to the newly formed cluster $C_{<k,l>}$, the log-likelihood $L$ decreases by

$$L_k(\theta_k) + L_l(\theta_l) - L_{<k,l>}(\theta_{<k,l>}) \equiv d(k,l) \geq 0 \quad (3)$$

Because $d(k, l)$ depends only on the cases belonging to the clusters involved in the merge, all the distances $d(k', l')$ between other cluster pairs remain unchanged. Moreover, the ML parameter set $\theta_k$ and the value of $d(k, l)$ depend only on a set of sufficient statistics that are stored for each cluster; and these items can be recursively updated when two clusters are merged without direct access to the cases. Using this fact, we have developed a memory efficient distance updating algorithm, described in Meilă and Heckerman (1998).

Our HAC implementation requires $\mathcal{O}(N)$ memory and time between $\mathcal{O}(N^2)$ and $\mathcal{O}(N^3)$. Experiments (section 6.2) show that typically the running time is close to the lower bound $\mathcal{O}(N^2)$. The algorithm can be generalized to any distance measure that is *local* (i.e. $d(k, l)$ depends only on $C_k$ and $C_l$) and, with minor modifications, also to certain classes of Gaussian mixtures. Finally, note that, whereas the EM and CEM require an initial point in the parameter space, HAC does not.

## 2.3 Initialization methods

As we have discussed, the EM and CEM algorithms require an initial set of parameter values. In this section, we describe the initialization methods that we compare.

First, however, we caution that "the best initialization method" is an ill-defined notion, becuase there is no formal delimitation between initial search and search. For example, when considering the EM algorithm, the ideal initial point would lie somewhere in the domain of attraction of the global optimum. Finding such a point, however, means finding the global optimum to a certain accuracy. This task represents a search per se, and perhaps a more challenging one than that performed by the main EM algorithm. Furthermore, the notion of "best" will generally involve a tradeoff between accuracy and computation cost. For these reasons, we should not expect to find an initialization method that outperforms all the others on all tasks. Rather, the performance (and relative performance) of an initialization procedure will depend on the data, the accuracy/cost tradeoff, and the "main" search algorithm. As is the case for search algorithms, one can only hope to find initialization methods that perform well on limited classes of tasks that arise in practice. This is the aim of our study on initialization methods.

The first method, the Random approach, consists of initializing the parameters of the model independently of the data. Using this approach, we sample the parameter values from an uninformative distribution. In our experiments, we sample the parameters of $P(X_i|k)$ from a uniform Dirichlet distribution with an equivalent sample size equal to the number of states of $X_i$.

The noisy-marginal method of Thiesson, Meek, Chickering, & Heckerman (1997) (herein denoted "Marginal") is a data dependent initialization method. Using this approach, we first determine the ML (or MAP) parameter configuration under the assumption that there is only one class. This step can be done in closed form. Next, for each class $k$ and each variable $X_i$, we create a conjugate (Dirichlet) distribution for the parameters corresponding to $P(X_i|k)$ whose parameter configuration of maximum value agrees with the ML or MAP configuration just computed and whose sample size is specified by the user. We then sample the parameters corresponding to $P(X_i|k)$ from this distribution. In our experiments, we use an equivalent sample size of two and match to the MAP configuration (given a uniform parameter prior).

The distribution of the hidden class variable is initialized to be the uniform distribution when using either of the above methods.

The last initialization method is hierarchical agglomerative clustering itself. In this data-dependent method, we perform HAC on a random sample of the data. From the resulting clusters, we extract a set of sufficient statistics (counts). We then set the model parameters to be the MAP given the counts obtained from agglomeration and a uniform (Dirichlet) prior distribution.

We apply agglomeration to a random sample of the data, because using all the data is often intractable. Furthermore, inaccuracies due to subsampling may be corrected by the EM and CEM clustering algorithms.

### 2.4 Performance criteria

In this section, we describe the criteria that we use to compare learned models. Some of our criteria measure the quality of the entire model (parameters and structure), whereas other criteria measure the quality of only the model structure—that is, the quality of the assumption that the data is generated from a mixture model with $K$ components.

As we have discussed, the log-marginal-likelihood criterion is used to select the best model structure (best number of clusters). We use this score in our comparisons as well.

Another criterion for model structure is the difference between the number of clusters in the model and the true number of clusters $K^{true}$. Such a measure reflects (in part) how well the learned models help the user to understand the domain under study. This criterion can only be determined for synthetic datasets where the true number of clusters is available. Furthermore, even when the true model available, this criterion may be inaccurate for small datasets, when there is insufficient data to support the true number of clusters. We note that, under certain experimental conditions, some of the learned clusters can be quite small. In the results that we present, we discard (i.e., do not include in the count of number of clusters) any cluster that has less than one case as its member. This situation can arise in models learned by the EM algorithm, due to the fractional assignment of cases to clusters.

A criterion for the entire model is the cross entropy between the true joint distribution for $X$ and the joint distribution given by the model:

$$\sum_x P^{true}(x) \log P(x|model) \qquad (4)$$

where *model* denotes both the parameters and structure of the model. In our experiments, we estimate this criterion using a holdout dataset $D_{test}$:

$$L_{test} = \frac{1}{|D_{test}|} \sum_{x \in D_{test}} \log P(x|model) \qquad (5)$$

Another criterion for the entire model is *classification accuracy*, defined to be the proportion of cases for which the most likely class $k$ (according to the model) is the true one. We determine the classification accuracy as follows. Because clusters in a learned model are interchangeable, we first must map the learned cluster labels to the actual ones. To do so, we construct the confusion matrix $C$:

$$C_{ii'} = \#\text{cases } j \text{ for which } k^j = i, k^{j*} = i' \qquad (6)$$

where $j$ is the case number, $k^j$ is the class that the learned model assigns to case $j$, and $k^{j*}$ is the true class of case $j$. Then, we map each cluster $k$ in the learned model to the cluster $k'$ of the true model to which most of its cases belong (this corresponds to a permutation of the rows of $C$). Once we have mapped the cluster labels, we simply count the cases in $k$ that have the correct label, and divide by the total number of cases. This criterion can be computed only for synthetic datasets where the true class information is available.

Practical criteria for algorithm comparison include running time and memory requirements. Because all the algorithms that we consider require memory proportional to the number of cases $N$, the number of observable variables $n$, and the number of clusters $K$, there is no further need to make experimental comparison with respect to storage usage. Nonetheless, running times per iteration differ for the considered algorithms. Moreover, the number of iterations to convergence for EM and CEM is a factor that cannot be predicted. Hence, experimental comparisons with respect to running time should be informative.

A summary of the performance criteria is given in Table 1. In the results that we report, we normalize the marginal likelihood and holdout scores by dividing the number of cases in the appropriate dataset. Also, we use base-two logarithms. Hence, both likelihoods are measured in bits per case.

## 3 Experimental procedure

An experimental condition is defined to be a choice of clustering algorithm, initialization algorithm, and parameters for each (e.g., the convergence criterion for EM and the number of subsamples for HAC). We evaluate each experimental condition as follows. First, we learn a sequence of models $model_{K_{min}}, \ldots model_{K_{max}}$ using the training set $D_{train}$. For each $K$, the model structure of $model_K$ is evaluated using the log-marginal-likelihood criterion (in particular, the Cheeseman-Stutz appoximation). Then, the number of clusters $K^*$ is chosen to be

$$K^* = \operatorname*{argmax}_{K} \log P(D_{train}|K) \qquad (7)$$

Once $K^*$ is selected, the corresponding trained model is evaluated using all criteria.

Because the quality of learned models is vulnerable to noise—randomness in the initial set of parameters for EM and CEM, and the subsample of points used for HAC—we repeat each experimental condition several times using different random seeds. We call an evaluation for a given experimental condition and random seed a "run".

We compute MAP parameters (as opposed to ML parameters) for the EM and CEM algorithms, using an uniform prior for each parameter. For HAC, we use ML-based distance. In all trials (except those to be noted), we run EM and CEM until either the relative difference between successive values for the log posterior parameter probability is less than $10^{-6}$ or 150 iterations are reached.

All experiments are run on a P6 200 MHz computer.

## 4 Datasets

### 4.1 The synthetic dataset

Synthetic datasets have the advantage that all of our criteria can be used to compare the clustering and intialization algorithms. Of course, one disadvantage of using such datasets is that the comparisons may not be realistic. To help overcome this concern, we constructed a synthetic model so as to mimic (as much as we could determine) a real-world dataset.

The real-world data set that served as the template for our synthetic model was obtained from the MSNBC news service. The dataset is a record of the stories read and not read by users of the www.msnbc.com web site during a one-week period in October of 1998. In this dataset, each observable variable corresponds to a story and has two states: "hit" (read) and "not hit" (not read). We shall use $X_i$ to refer both to a particular story and its corresponding variable.

A preliminary clustering analysis of this dataset, using both EM and CEM with random initialization, showed the following. (1) There were approximately 10 clusters. (2) The size of clusters followed Zipf's law (Zipf, 1949). That is, the probabilities $P(\text{class} = k), k = 1, \ldots, K$, when sorted in descending order, showed a power-law decay. (3) The marginal probabilities of story hits also followed Zipf's law. That is, the probabilities $P(X_i =\text{hit})$ for all stories, when sorted in descending order, showed a power-law decay. (4) The clusters overlapped. That is, many users had substantial class membership in more than one cluster. (5) Users in smaller clusters tended to hit more stories. (6) The clusters obtained did not vary significantly when all but the 150 most commonly hit stories were discarded from the analysis. This finding is likely due to item 3.

We used all of these observations in the construction of the synthetic model. In addition, we wanted the synthetic model to be more complex than the models we would attempt to learn (the naive-Bayes model). Consequently, we constructed the model as follows. First, we built a naive-Bayes model where the hidden variable class had $K = 10$ states and where the observable variables corresponded to the 300 most commonly hit stories. We assigned the distribution $P(\text{class})$ to be (0.25, 0.18, 0.18, 0.09, 0.09, 0.09, 0.045, 0.035, 0.025, 0.015)–roughly approximating a power decay. Then, for each story variable $X_i$ and for $k = 1, \ldots, 10$, we assigned $P(X_i =\text{hit}|\text{class} = k)$ to be the marginal distribution for story $X_i$, and perturbed these conditional distributions with noise to separate the clusters. In particular, for every $X_i$ and for $k = 1, \ldots, 10$, we perturbed the log odds $\log P(X_i = \text{hit}|\text{class} =$ $k)/(1 - P(X_i = \text{hit}|\text{class} = k))$ with normal noise $N(\alpha, 1)$, where $\alpha = -0.5$ for class $= 1, 2, 3$ (the large clusters), $\alpha = 0$ for class $= 4, 5, 6$ (the medium-size clusters), and $\alpha = 1$ for class $= 7, 8, 9, 10$ (the small clusters). These values for $\alpha$ produced a model with overlapping clusters such that smaller clusters contained more hits. Next, we added arcs between the observable variables such that each observed variable had an average of two parents and a maximum of three parents. To parameterize the conditional dependencies among the observed variables, we perturbed the log odds $\log P(X_i = \text{hit}|Pa_i, \text{class} = k)/(1 - P(X_i = \text{hit}|Pa_i, \text{class} = k))$ for every parent configuration with normal noise $N(0, 0.25)$.

Table 1: Performance criteria.

| Criterion | Expression | Comment |
|---|---|---|
| Marginal L | $\frac{1}{|D_{train}|} \log_2 P(D_{train}|K^*)$ | Bayesian criterion |
| $K^*$ | $K^*$ | number of clusters in "best" model |
| Class acc | $\frac{1}{|D_{test}|}$ #cases correctly classified | classification accuracy |
| Holdout L | $\frac{1}{|D_{test}|} \log_2 P(D_{test}|model_{K^*})$ | prediction accuracy on a test set |
| Runtime | | training time for $model_{K^*}$ |

Finally, we sampled 32,000 cases from the model and then discarded the 150 least commonly hit stories. By discarding these variables (i.e., making them unobserved), we introduced additional dependencies among the remaining observed variables. We refer to this data set as $SY_{32K}$. We also generated a test set $SY_{8K}$ containing 8,000 cases for the same 150 variables retained in $SY_{32K}$. The generative model and datasets are available via anonymous ftp at ftp.research.microsoft.com/pub/dtg/msnbc-syn.

### 4.2 The digits datasets

Another source of data for our comparison consists of images of handwritten digits made available by the US Postal Service Office for Advanced Technology (Frey, Hinton, & Dayan, 1995). The images were normalized and quantized to 8x8 dimensional binary patterns. For each digit we had a training set of 700 cases and a test set of 400 cases. We present detailed results on one digit, namely **digit6**. Results for other digits that we examined ("0" and "2") are similar.

## 5 Comparison of clustering algorithms

This section presents a comparison between the EM, CEM, and HAC algorithms on the **synthetic** and **digit6** datasets. Because HAC was too slow to run on a full 32K training set (for synthetic data), we experimented with HAC on a subset $SY'_{8K}$ of $SY_{32K}$ with only 8,000 cases. However, this training set was too small to learn a model with the complexity of the true one. Hence, to provide a fair comparison, we first compared EM and CEM using the full $SY_{32K}$ dataset, and then compared the better of these two algorithms with HAC using $SY'_{8K}$.

### 5.1 Synthetic data: EM versus CEM

As mentioned above, for the purpose of this comparison, $D_{train} = SY_{32K}$ and $D_{test} = SY_{8K}$. Because both EM and CEM require initial parameters, we compared these algorithms using all three initialization methods.

The results on the **synthetic** data are presented in Table 2. In this and subsequent tables, boldface is used to indicate the best algorithm for each criterion. The table shows the clear dominance of EM over CEM for all initialization methods and for all criteria. The most striking difference is in the choice of $K^*$, the number of clusters. CEM constantly underestimated $K^*$, whereas EM, for the two out of three initialization methods used, successfully found the true number of clusters.

Several issues concerning classification accuracy are worth noting. First, all classification accuracies were low, because the clusters overlapped significantly. In particular, the classification accuracy for the true model was only 73%. Also, classification accuracy correlated closely with the choice of $K^*$. When $K^* = 10$, the classification accuracy of the learned model was close to that of the true model. Whereas, when $K^* = 7$ or lower, the classification accuracy of the learned model was approximately two-thirds of that of the true model. ¿From an examination of the confusion matrices for CEM, we found that the underestimation of $K^*$ had two sources: confusion among the 3 largest clusters, and the failure to discover the existence of the smallest clusters. The second source had a smaller impact on classification accuracy.

The only possible advantage of CEM over EM is running time. CEM was about four times faster than EM, because (1) EM requires the accumulation of fractional statistics whereas CEM does not, and (2) CEM takes fewer iterations to converge. We attribute the second phenomenon to the fact that CEM's explorations were more constrained, so that the impossibility of an upward move occurred earlier. We shall further examine this issue in Section 5.4.

Table 3: Performance of the EM and HAC algorithms on the **synthetic** dataset (average and standard deviation over five runs). Runtimes are reported in minutes per class.

|  | **EM** | **HAC** |
|---|---|---|
| Marginal L | **-20.93 ± 0.108** | -21.08 ± 0.144 |
| $K^*$ | **4 ± 0** | 2 ± 1 |
| Holdout L | **-20.63 ± 0.018** | -20.88 ± 0.018 |
| Class acc | **0.41 ± 0.01** | 0.33 ± 0.02 |
| Runtime | **0.6**; $\mathcal{O}(N)$ | 35; $\mathcal{O}(N^2)$ |

### 5.2 Synthetic data: EM versus HAC

As discussed, we next compared EM—the better of EM and CEM—with the HAC algorithm using a smaller training set $D_{train} = SY'_{8K}$. In this comparison, we used the same test set $D_{test} = SY_{8K}$. EM was initialized with the Marginal method.

The results in Table 3 show the clear superiority of EM over HAC. For such a small dataset, both algorithms fare poorly in terms of the number of clusters found, but EM finds twice as many. Indeed, an analysis of the confusion matrices showed that HAC was unable to distinguish the three largest clusters. In addition, an important difference is seen in the running time. HAC runs approximately 60 times slower than EM; and this ratio grows with $N$ because the running times of EM and HAC are $\mathcal{O}(N)$ and approximately $\mathcal{O}(N^2)$, respectively.

### 5.3 Digits data: Comparison of EM, CEM, and HAC

For the **digit6** dataset, all three algorithms were trained (tested) on the same training (test) set. The results, showing marginal-likelihood and holdout scores for **digit6** are given in Table 4. Only the results for Marginal intialization of the EM and CEM algorithms are shown. The results for the other intialization methods are similar.

The EM algorithm performed best by both criteria. We note that, for this dataset, all the methods choose about the same number of clusters.

### 5.4 EM versus CEM: Runtime considerations

On our datasets, the EM algorithm dominates HAC, because EM is both more accurate and more efficient.

Table 4: Performance of the EM, CEM, and HAC algorithms on the **digit6** dataset (averaged over 20 runs for EM and 10 runs for CEM). EM and CEM were both initialized by the Marginal method.

|  | **EM** | **CEM** | **HAC** |
|---|---|---|---|
| Marginal L | **-35.09 ± 0.04** | -35.30 ± 0.07 | -35.76 |
| Holdout L | **-32.36 ± 0.22** | -32.97 ± 0.30 | -32.42 |

Table 5: Performance of the EM and CEM algorithms on the **synthetic** dataset when EM is allowed to run no longer than CEM (averaged over five runs). Runtimes are reported in seconds per class.

|  | **EM** | **CEM** |
|---|---|---|
| Marginal L | **-20.57 ± 0.026** | -20.66 ± 0.031 |
| $K^*$ | **9 ± 2** | 7 ± 3 |
| Holdout L | **-20.42 ± 0.032** | -20.52 ± 0.036 |
| Class acc | **0.57 ± 0.03** | 0.46 ± 0.06 |
| Runtime | **25** | 27 |

So far, however, the case for EM versus CEM is not so clear. As we have seen, the EM algorithm is more accurate, but less efficient. This brings us to the question: If EM is forced to run for a time shorter or equal to the time taken by CEM is it still more accurate than CEM? In this section, we examine this question.

We adjusted the running time of EM by changing the convergence threshold. We conducted timing experiments and found that convergence thresholds of $4 \times 10^{-4}$ and $10^{-3}$ (for the synthetic and digits data sets, respectively) yielded EM speeds that were slightly faster than CEM. We repeated the comparison of EM and CEM, using these new thresholds.

The results for the synthetic and digits datasets are shown in Tables 5 and 6, respectively. For both datasets and all criteria, EM is still more accurate than CEM, although the differences in accuracy are less than what we obtained for the original conver-gence threshold. All experimental conditions yield a significant difference at the 95% level, except the $K^*$ criterion in the synthetic dataset and the marginal-likelihood criterion in the digits dataset which are significant only at the 85% confidence level.

Table 6: Performance of the EM and CEM algorithms on the **digit6** dataset when the convergence criterion for EM is set so that EM runs slightly faster than CEM (averaged over ten runs). Runtimes are reported in seconds per class.

|  | **EM** | **CEM** |
|---|---|---|
| Marginal L | **-35.26 ± 0.09** | **-35.30 ± 0.07** |
| Holdout L | **-32.47 ± 0.23** | -32.97 ± 0.03 |
| Runtime | **0.082** | 0.089 |

Table 2: Performance of the EM and CEM algorithms with three different initialization methods on the **synthetic** dataset (average and standard deviation over five runs). The classification accuracy of the true model is 0.73. Runtimes are reported in minutes per class and exclude initialization. Boldface is used to indicate the best algorithm for each criterion.

| **Initialization** | Random | | Marginal | | **HAC** | |
|---|---|---|---|---|---|---|
| | **EM** | **CEM** | **EM** | **CEM** | **EM** | **CEM** |
| Marginal L (bits/case) | **-20.53** ±**0.009** | -20.57 ±0.032 | **-20.51** ±**0.0045** | -20.66 ±0.031 | **-20.52** ±**0.013** | -20.72 ±0.031 |
| $K^*$ | 7±1 | 6±3 | 10±1 | 7±3 | 10±1 | 7±3 |
| Holdout L (bits/case) | **-20.41** ±**0.036** | -20.50 ±0.036 | **-20.36** ±**0.018** | -20.52 ±0.036 | **-20.34** ±**0.018** | -20.72 ±0.072 |
| Class acc | **0.50±0.05** | 0.44±0.01 | **0.66±0.04** | 0.46±0.06 | **0.68±0.01** | 0.43±0.04 |
| Runtime | 2 | **0.5** | 2 | **0.5** | 2 | **0.5** |

## 6 Comparison of initialization methods

We now examine the influence of initialization procedures on the performance of the EM algorithm—the algorithm that performed best in our previous comparison. Our main concern will be the quality of the models obtained. Running time, as long as it is in reasonable limits, will be of secondary interest.

### 6.1 Synthetic data

The results on the synthetic dataset were shown previously in Table 2. We used $D_{train} = SY_{32K}$ and $D_{test} = SY_{8K}$. We used a subsample of the training set of size $N' = 2000$ for the HAC intialization method.

The data independent (Random) method fared worse than did the data dependent methods across all criteria. All differences between the data dependent methods and Random—except one: HAC versus Random on Marginal Likelihood—are significant. On the other hand, there are no significant differences between Marginal and HAC.

The initialization runtimes for Random, Marginal, and HAC were 0, 4, and 1800 seconds, respectively. Random is fastest (taking constant time), Marginal is slightly slower (requiring one sweep through the dataset), and HAC is hundreds of times slower yet, even though only a portion of the training data was used.

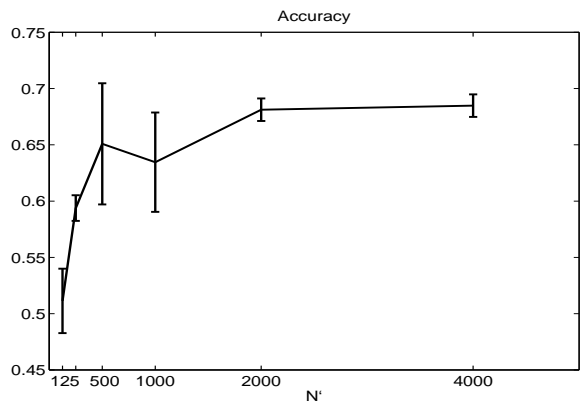

Figure 2: Classification accuracy of the final EM solution versus the HAC subsample size $N'$ for the **synthetic** data. Statistics are computed over five runs.

### 6.2 Sample size for hierarchical agglomerative clustering

In the previous section, we saw that HAC used on a sample of the original training set of size $N'$, can provide a good set of initial parameters for the EM algorithm. In this section, we ask the question: How small can we make $N'$ and still obtain good performance?

To answer this question, we varied $N'$ from 125 to 4000. Figure 2 shows the classification accuracy of the resulting models. We obtained similar curves for the other performance.

The graph shows that, as $N'$ decreases from 4000 to 500, the performance decays slowly, accompanied by an increasing variability. Not surprising, an analysis of the confusion matrices showed that performance decreases because, as $N'$ decreases, the smaller clusters are missed and the confusion between the larger ones increase. These observations suggest that the lower limit on $N'$ should be influenced by prior knowledge about the size of the smallest cluster. For $N'$ below 500, the classification performance degrades abruptly. Detailed analysis showed that, in this range, the clustering algorithm was failing to separate the largest clusters.

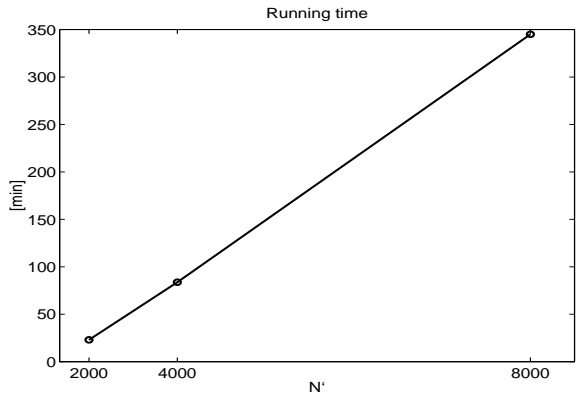

Figure 3: Running time of the EM algorithm initialized with HAC versus the HAC subsample size $N'$. The error bars are small and have been omitted. The scale for $N'$ is quadratic.

The running time of HAC versus the number of samples is shown in Figure 3. In this range for $N'$, the time taken by HAC initialization strongly dominates the running time for the EM algorithm. Although the theoretical worst case is $\mathcal{O}(N^3)$, the graph shows that the running time approximately follows a quadratic law.

### 6.3 The digits6 dataset

We compare the three initialization methods on the **digit6** data as well. We examine HAC with $N' = N = 700$ and with lower values $N' = 100, 300$. Table 7 summarizes the results. There are no clear winners. In contrast to the results for the synthetic data set, Random does not perform worse. Nonetheless, we find two related qualitative differences between the models produced by Random initialization and the other two methods. The clusters learned using Random initialization are greater in number and show more variability in size. Moreover, Random initialization tends to yield more small clusters. On average, for the models whose performance is recorded in Table 7, 1.5 clusters are supported by less than one case.

Finally, note that HAC followed by EM produces better models than does HAC alone (see Tables 4 and 7).

## 7 Discussion

We have compared several popular batch algorithms for clustering in high dimensions on discrete-variable models with uninformative priors. To do so, we have formulated a likelihood based disatance measure for agglomerative clustering over discrete variable domains and have introduced a new, memory efficient HAC algorithm. Although comparisons with additional data sets (including ones with larger dimension and continuous variables) are needed, our results suggest that the EM algorithm produces better models than does CEM and HAC, regardless of the criterion for model evaluation. We found that, for original convergence settings, EM was slower than CEM and HAC was slower yet. Nonetheless, we found that the convergence setting of EM could be adjusted so that EM ran faster than CEM and still produced better models, albeit of lower quality than for the original settings.

These results suggest that the quality of a clustering algorithm correlates well with assignment "softness". Namely, HAC assigns each case to only one cluster and this assignment cannot be changed in subsequent iterations. CEM also assigns a case to only one cluster, but each iteration recomputes the assignment. EM not only reevaluates its assignments at each iteration (like CEM), but also allows for partial credit to several clusters.

Finding that the EM algorithm performed best, we studied various ways of choosing initial parameters. Although the three methods used are dissimilar, their performance is strikingly similar on all the datasets we examined. On the synthetic dataset, the Random method performed worse than did the data-dependent initialization method, but this difference was not seen for the real-world data. Because we found no difference between HAC (as an initialization method) and Marginal, except that Marginal is more efficient, our results suggest that that Marginal initialization is the method of choice. Finding situations when Random and Marginal produce different results is a possible topic for further research.

Although our results are suggestive, comparisons using additional datasets—for example, ones having 200 or more dimensions and ones that contain continuous measurements—are needed. In addition, comparisons with more sophisticated variants of the EM algorithm such as EM with conjugate-gradient acceleration (Thiesson, 1995) and the EM$\nu$ algorithm (Bauer, Koller, & Singer, 1997) should be performed.

Table 7: Performance of the EM algorithm when initialized by the Random, Marginal, and HAC methods on the **digit6** dataset (average and standard deviation over 12 or more runs).

| Initialization | Random | Marginal | HAC $N' = 100$ | HAC $N' = 300$ | HAC $N' = N = 700$ |
|---|---|---|---|---|---|
| Marginal L (bits/case) | -35.063 ±0.040 | -35.089 ±0.042 | -35.091 ± 0.047 | -35.087 ±0.042 | -35.073 |
| Holdout L (bits/case) | -32.53 ±0.255 | -32.36 ±0.224 | -32.04 ±0.513 | -32.18 ±0.238 | -32.27 |

## Acknowledgments

We thank Max Chickering, Chris Meek, and Bo Thiesson for their assistance with the implementation of the algorithms and for many useful and interesting discussions. We also thank Steven White and Ian Marriott for providing the original MSNBC dataset.